\documentclass[final]{cvpr}

\usepackage{times}
\usepackage{epsfig}
\usepackage{graphicx}
\usepackage{amsmath}
\usepackage{amssymb}

\usepackage{multirow}
\usepackage{bm}
\usepackage{algorithm}
\usepackage{algorithmicx}
\usepackage{algpseudocode}
\usepackage{color}
\usepackage{enumitem}
\usepackage{arydshln}
\usepackage{pifont}
\usepackage{tablefootnote}
\usepackage{subfigure}
\usepackage{microtype}

\usepackage[pagebackref=true,breaklinks=true,colorlinks,bookmarks=false]{hyperref}

\def\cvprPaperID{5142} 
\def\confYear{CVPR 2021}

\begin{document}

\title{Read Like Humans: Autonomous, Bidirectional and Iterative Language Modeling for Scene Text Recognition}

\author{Shancheng Fang \quad Hongtao Xie\thanks{The corresponding author} \quad Yuxin Wang \quad Zhendong Mao \quad Yongdong Zhang \\
University of Science and Technology of China\\
{\tt\small  \{fangsc, htxie, zdmao, zhyd73\}@ustc.edu.cn, wangyx58@mail.ustc.edu.cn}
}

\maketitle

\begin{abstract}

Linguistic knowledge is of great benefit to scene text recognition. However, how to effectively model linguistic rules in end-to-end deep networks remains a research challenge. In this paper, we argue that the limited capacity of language models comes from: 1) implicitly language modeling; 2) unidirectional feature representation; and 3) language model with noise input. Correspondingly, we propose an autonomous, bidirectional and iterative ABINet for scene text recognition. Firstly, the autonomous suggests to block gradient flow between vision and language models to enforce explicitly language modeling. Secondly, a novel bidirectional cloze network (BCN) as the language model is proposed based on bidirectional feature representation. Thirdly, we propose an execution manner of iterative correction for language model which can effectively alleviate the impact of noise input. Additionally, based on the ensemble of iterative predictions, we propose a self-training method which can learn from unlabeled images effectively. Extensive experiments indicate that ABINet has superiority on low-quality images and achieves state-of-the-art results on several mainstream benchmarks. Besides, the ABINet trained with ensemble self-training shows promising improvement in realizing human-level recognition. Code is available at \url{https://github.com/FangShancheng/ABINet}.

\end{abstract}

\section{Introduction}

Possessing the capability of reading text from scene images is indispensable to artificial intelligence~\cite{long2020scene,wan2020vocabulary}. To this end, early attempts regard characters as meaningless symbols and recognize the symbols by classification models~\cite{wang2011end, jaderberg2016reading}. However, when confronted with challenging environments such as occlusion, blur, noise, etc., it becomes faint due to out of visual discrimination. Fortunately, as text carries rich linguistic information, characters can be reasoned according to the context. Therefore, a bunch of methods~\cite{jaderberg2014deep,jaderberg2015deep,qiao2020seed} turn their attention to language modeling and achieve undoubted improvement.

However, how to effectively model the linguistic behavior in human reading is still an open problem. From the observations of psychology, we can make three assumptions about human reading that language modeling is autonomous, bidirectional and iterative: 1) as both deaf-mute and blind people could have fully functional vision and language separately, we use the term \emph{autonomous} to interpret the independence of learning between vision and language. The \emph{autonomous} also implies a good interaction between vision and language that independently learned language knowledge could contribute to the recognition of characters in vision. 2) The action of reasoning character context behaves like cloze task since illegible characters can be viewed as blanks. Thus, prediction can be made using the cues of legible characters on the left side and right side of the illegible characters simultaneously, which is corresponding to the \emph{bidirectional}. 3) The \emph{iterative} describes that under the challenging environments, humans adopt a progressive strategy to improve prediction confidence by iteratively correcting the recognized results.

\begin{figure}
   \begin{center}
      \includegraphics[width=0.48\textwidth]{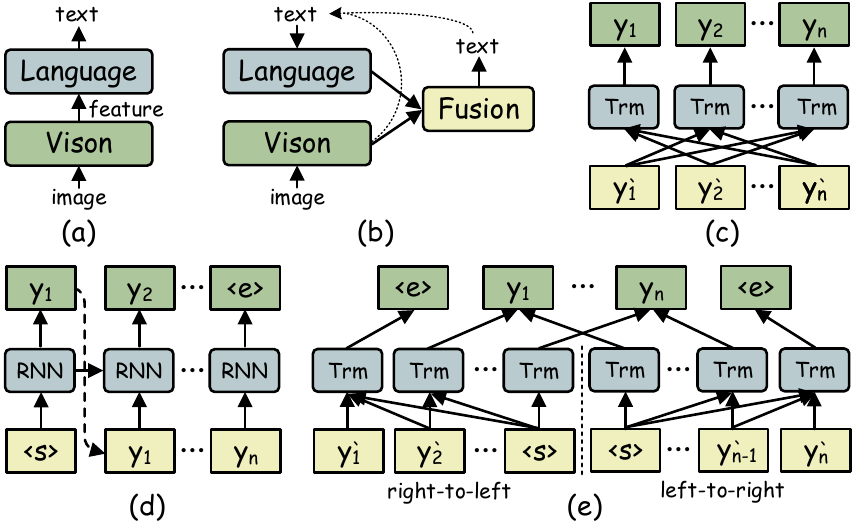}
      \caption{(a) Coupled language model. (b) Our autonomous language model with iterative correction. (c) Our bidirectional structure. (d) Unidirectional RNN in auto-regression. (e) Ensemble of two unidirectional Transformers in parallel-prediction.}
      \label{fig:overall}
   \end{center}
   \vspace{-2.5em}
\end{figure}

Firstly, applying the \textbf{autonomous} principle to scene text recognition~(STR) means that recognition models should be decoupled into vision model~(VM) and language model~(LM), and the sub-models could be served as functional units independently and learned separately. Recent attention-based methods typically design LMs based on RNNs or Transformer~\cite{vaswani2017attention}, where the linguistic rules are learned \emph{implicitly} within a coupled model~\cite{lee2016recursive, shi2018aster, sheng2019nrtr} (Fig.~\ref{fig:overall}a). Nevertheless, whether and how well the LMs learn character relationship is unknowable. Besides, this kind of methods is infeasible to capture rich prior knowledge by directly pre-training LM from large-scale unlabeled text.

Secondly, compared with the unidirectional LMs~\cite{sundermeyer2012lstm}, LMs with \textbf{bidirectional} principle capture twice the amount of information. A straightforward way to construct a bidirectional model is to merge a left-to-right model and a right-to-left model~\cite{peters2018deep, devlin2018bert}, either in probability-level~\cite{wang2020decoupled,shi2018aster} or in feature-level~\cite{yu2020towards} (Fig.~\ref{fig:overall}e). However, they are strictly less powerful as their language features are unidirectional \emph{representation} in fact. Also, the ensemble models mean twice as expensive both in computations and parameters. A recent striking work in NLP is BERT~\cite{devlin2018bert}, which introduces a deep bidirectional representation learned by masking text tokens. Directly applying BERT to STR requires masking all the characters within a text instance, whereas this is extremely expensive since each time only one character can be masked.

Thirdly, LMs executed with \textbf{iterative} principle can refine the prediction from visual and linguistic cues, which is not explored in current methods. The canonical way to perform an LM is auto-regression~\cite{wang2020decoupled,cheng2017focusing,wojna2017attention} (Fig.~\ref{fig:overall}d), in which error recognition is accumulated as noise and taken as input for the following prediction. To adapt the Transformer architectures, ~\cite{lyu20192d,yu2020towards} give up auto-regression and adopt parallel-prediction (Fig.~\ref{fig:overall}e) to improve efficiency. However, noise input still exists in parallel-prediction where errors from VM output directly harm the LM accuracy. In addition, parallel-prediction in SRN~\cite{yu2020towards} suffers from unaligned-length problem that SRN is tough to infer correct characters if text length is wrongly predicted by VM.

Considering the deficiencies of current methods from the aspects of internal interaction, feature representation and execution manner, we propose ABINet guided by the principles of \emph{Autonomous}, \emph{Bidirectional} and \emph{Iterative}. Firstly, we explore a decoupled method~(Fig.~\ref{fig:overall}b) by blocking gradient flow (BGF) between VM and LM, which enforces LM to learn linguistic rules explicitly. Besides, both VM and LM are autonomous units and could be pre-trained from images and text separately. Secondly, we design a novel bidirectional cloze network (BCN) as the LM, which eliminates the dilemma of combining two unidirectional models~(Fig.~\ref{fig:overall}c). The BCN is jointly conditioned on both left and right context, by specifying attention masks to control the accessing of both side characters. Also, accessing across steps is not allowed to prevent leaking information. Thirdly, we propose an execution manner of iterative correction for LM~(Fig.~\ref{fig:overall}b). By feeding the outputs of ABINet into LM repeatedly, predictions can be refined progressively and the unaligned-length problem could be alleviated to a certain extent. Additionally, treating the iterative predictions as an ensemble, a semi-supervised method is explored based on self-training, which exploits a new solution toward human-level recognition.

Contributions of this paper mainly include: 1) we propose autonomous, bidirectional and iterative principles to guide the design of LM in STR. Under these principles the LM is a functional unit, which is required to extract bidirectional representation and correct prediction iteratively. 2) A novel BCN is introduced, which estimates the probability distribution of characters like cloze tasks using bidirectional representation. 3) The proposed ABINet achieves state-of-the-art (SOTA) performance on mainstream benchmarks, and the ABINet trained with ensemble self-training shows promising improvement in realizing human-level recognition.

\begin{figure*}
   \begin{center}
      \includegraphics[width=1.0\textwidth]{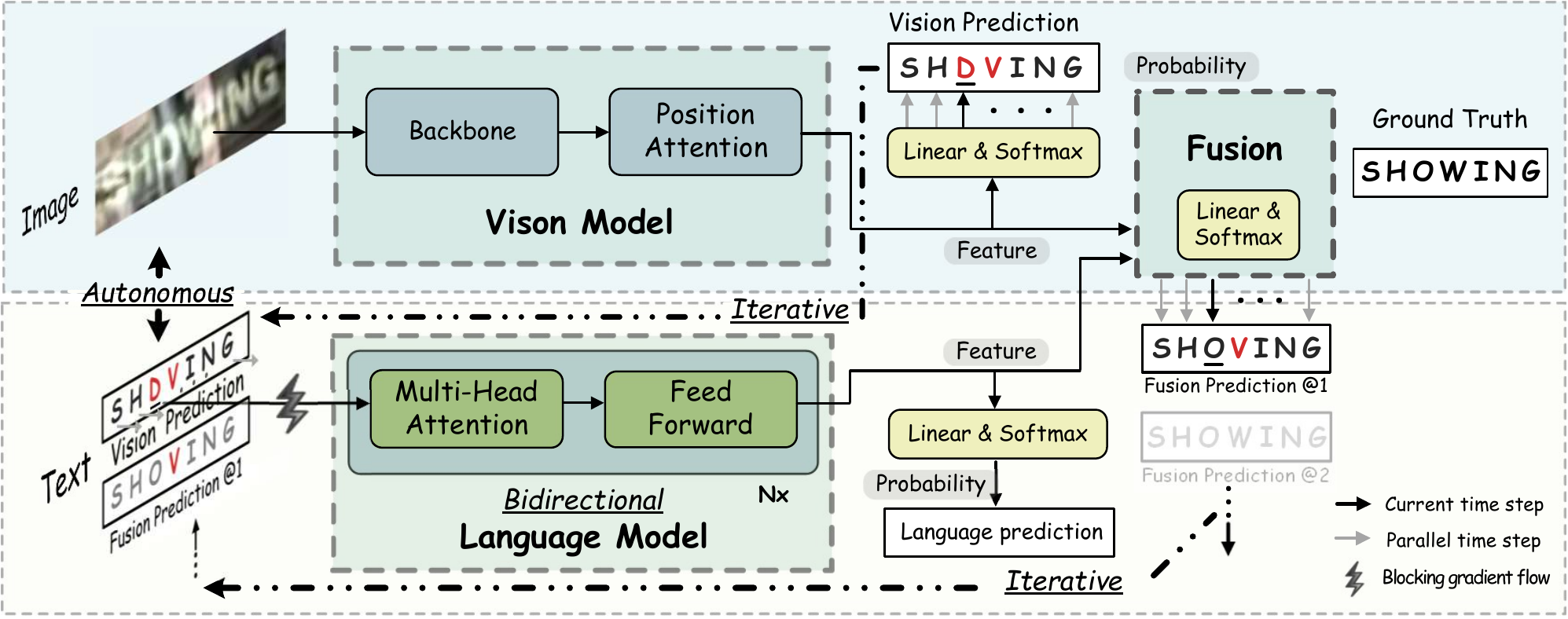}
      \caption{A schematic overview of ABINet.}
      \label{fig:framework}
   \end{center}
   \vspace{-2em}
 \end{figure*}

\section{Related Work}
\subsection{Language-free Methods} 

Language-free methods generally utilize visual features without the consideration of relationship between characters, such as CTC-based~\cite{graves2006connectionist} and segmentation-based~\cite{li2017fully} methods. The CTC-based methods employ CNN to extract visual features and RNN to model features sequence. Then the CNN and RNN are trained end-to-end using CTC loss~\cite{shi2016end,he2016reading,su2017accurate,hu2020gtc}. The segmentation-based methods apply FCN to segment characters in pixel-level. Liao~\etal recognize characters by grouping the segmented pixels into text regions. Wan~\etal~\cite{wan2019textscanner} propose an additional order segmentation map which transcripts characters in the correct order. Due to lacking of linguistic information, the language-free methods cannot resolve the recognition in low-quality images commendably.

\subsection{Language-based Methods}

\vspace{-0.5em}
\paragraph{Internal interaction between vision and language.} In some early works, bags of $N$-grams of text string are predicted by a CNN which acts as an explicit LM~\cite{jaderberg2015deep,jaderberg2014deep,jaderberg2014synthetic}. After that the attention-based methods become popular, which implicitly models language using more powerful RNN~\cite{lee2016recursive,shi2018aster} or Transformer~\cite{wang2019simple,sheng2019nrtr}. The attention-based methods follow encoder-decoder architecture, where the encoder processes images and the decoder generates characters by focusing on relevant information from 1D image features~\cite{lee2016recursive,shi2016robust,shi2018aster,cheng2017focusing,cheng2018aon} or 2D image features~\cite{yang2017learning,wojna2017attention,liao2019scene, li2019show}. For example, R$^2$AM~\cite{lee2016recursive} employs recursive CNN as a feature extractor and LSTM as a learned LM implicitly modeling language in character-level, which avoids the use of $N$-grams. Further, this kind of methods is usually boosted by integrating a rectification module~\cite{shi2018aster,zhan2019esir,yang2019symmetry} for irregular images before feeding the images into networks. Different from the methods above, our method strives to build a more powerful LM by explicitly language modeling. In attempting to improve the language expression, some works introduce multiple losses where an additional loss comes from semantics~\cite{qiao2020seed, lyu20192d, yu2020towards, fang2018attention}. Among them, SEED~\cite{qiao2020seed} proposes to use pre-trained FastText model to guide the training of RNN, which brings extra semantic information. We deviate from this as our method directly pre-trains LM in unlabeled text, which is more feasible in practice.

\vspace{-1.3em}
\paragraph{Representation of language features.} The character sequences in attention-based methods are generally modeled in left-to-right way~\cite{lee2016recursive, shi2016robust, cheng2017focusing, wan2019textscanner}. For instance, Textscanner~\cite{wan2019textscanner} inherits the unidirectional model of attention-based methods. Differently, they employ an additional position branch to enhance positional information and mitigate misrecogniton in contextless scenarios. To utilize bidirectional information, methods like~\cite{graves2008novel, shi2018aster, wang2020decoupled, yu2020towards} use an ensemble model of two unidirectional models. Specifically, to capture global semantic context, SRN~\cite{yu2020towards} combines features from a left-to-right and a right-to-left Transformers for further prediction. We emphasize that the ensemble bidirectional model is intrinsically a unidirectional feature representation.

\vspace{-1.3em}
\paragraph{Execution manner of language models.} Currently, the network architectures of LMs are mainly based on RNN and Transformer~\cite{vaswani2017attention}. The RNN-based LM is usually executed in auto-regression~\cite{wang2020decoupled,cheng2017focusing,wojna2017attention}, which takes the prediction of last character as input. Typical work such as DAN~\cite{wang2020decoupled} obtains the visual features of each character firstly using proposed convolutional alignment module. After that GRU predicts each character by taking the prediction embedding of the last time step and the character feature of the current time step as input. The Transformer-based methods have superiority in parallel execution, where the inputs of each time step are either visual features~\cite{lyu20192d} or character embedding from the prediction of visual feature~\cite{yu2020towards}. Our method falls into parallel execution, but we try to alleviate the issue of noise input existing in parallel language model.

\section{Proposed Method}
\subsection{Vision Model}

\begin{figure}
   \begin{center}
      \includegraphics[width=0.5\textwidth]{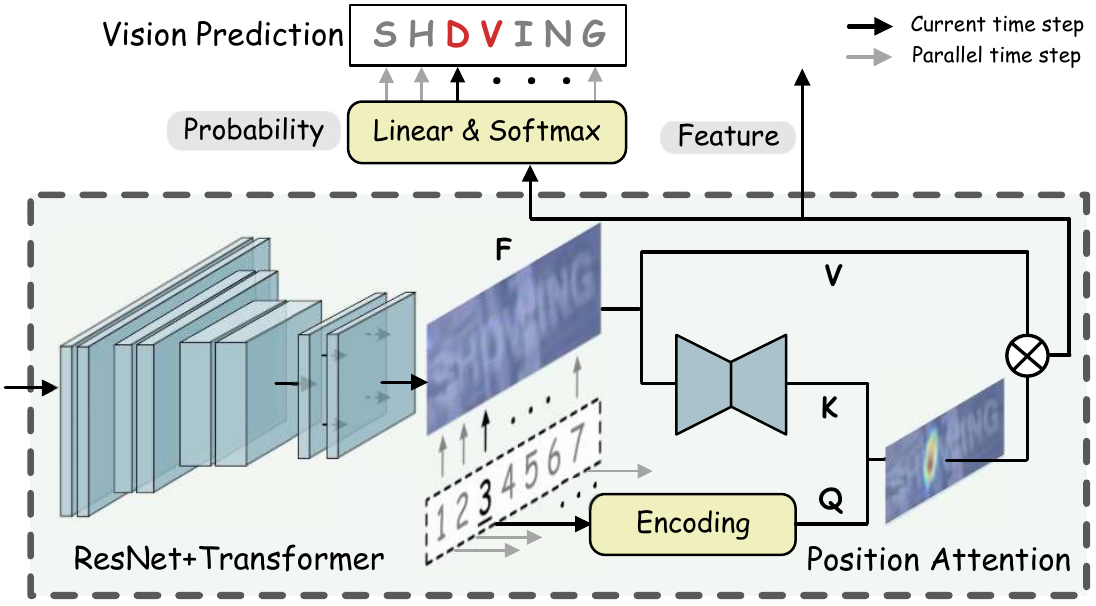}
      \caption{Architecture of vision model.}
      \label{fig:vision}
   \end{center}
   \vspace{-2em}
\end{figure}

The vision model consists of a backbone network and a position attention module (Fig.~\ref{fig:vision}). Following the previous methods, ResNet\footnote{There are 5 residual blocks in total and down-sampling is performed after the 1st and 3nd blocks.}~\cite{shi2018aster, wang2020decoupled} and Transformer units~\cite{yu2020towards, lyu20192d} are employed as the feature extraction network and the sequence modeling network. For image $\bm{x}$ we have:
\begin{equation}
\mathbf{F}_b = \mathcal{T}(\mathcal{R}(\bm{x})) \in \mathbb{R}^{\frac{H}{4} \times \frac{W}{4} \times C}, 
\end{equation}
where $H,W$ are the size of $\bm{x}$ and $C$ is feature dimension.

The module of position attention transcribes visual features into character probabilities in parallel, which is based on the query paradigm~\cite{vaswani2017attention}:
\begin{align}
\mathbf{F}_v = \text{softmax}(\frac{\mathbf{Q}\mathbf{K}^\mathsf{T}}{\sqrt{C}})\mathbf{V}.
\end{align}
Concretely, $\mathbf{Q} \in \mathbb{R}^{T \times C}$ is positional encodings~\cite{vaswani2017attention} of character orders and $T$ is the length of character sequence. $\mathbf{K} = \mathcal{G}(\mathbf{F}_b) \in \mathbb{R}^{\frac{HW}{16} \times C}$, where $\mathcal{G}(\cdot)$ is implemented by a mini U-Net\footnote{A network with 4-layer encoder, 64 channels, $add$ fusion and interpolation upsample.}~\cite{ronneberger2015u}. $\mathbf{V} = \mathcal{H}(\mathbf{F}_b) \in \mathbb{R}^{\frac{HW}{16} \times C}$, where $\mathcal{H}(\cdot)$ is identity mapping.

\subsection{Language Model}
\subsubsection{Autonomous Strategy}
\label{sec:autonomous}

As shown in Fig.~\ref{fig:framework}, the autonomous strategy includes following characteristics: 1) the LM is regarded as an independent model of spelling correction which takes probability vectors of characters as input and outputs probability distributions of expected characters. 2) The flow of training gradient is blocked (BGF) at input vectors. 3) The LM could be trained separately from unlabeled text data.

Following the strategy of autonomous, the ABINet can be divided into interpretable units. By taking the probability as input, LM could be replaceable (\ie, replaced with more powerful model directly) and flexible (\eg, executed iteratively in Section~\ref{sec:iterative}). Besides, an important point is that BGF enforces model to learn linguistic knowledge inevitably, which is radically distinguished from implicitly modeling where what the models exactly learn is unknowable. Furthermore, the autonomous strategy allows us to directly share the advanced progresses in NLP community. For instance, pre-training the LM can be an effective way to boost the performance.

\subsubsection{Bidirectional Representation}

\begin{figure}
   \begin{center}
      \includegraphics[width=0.4\textwidth]{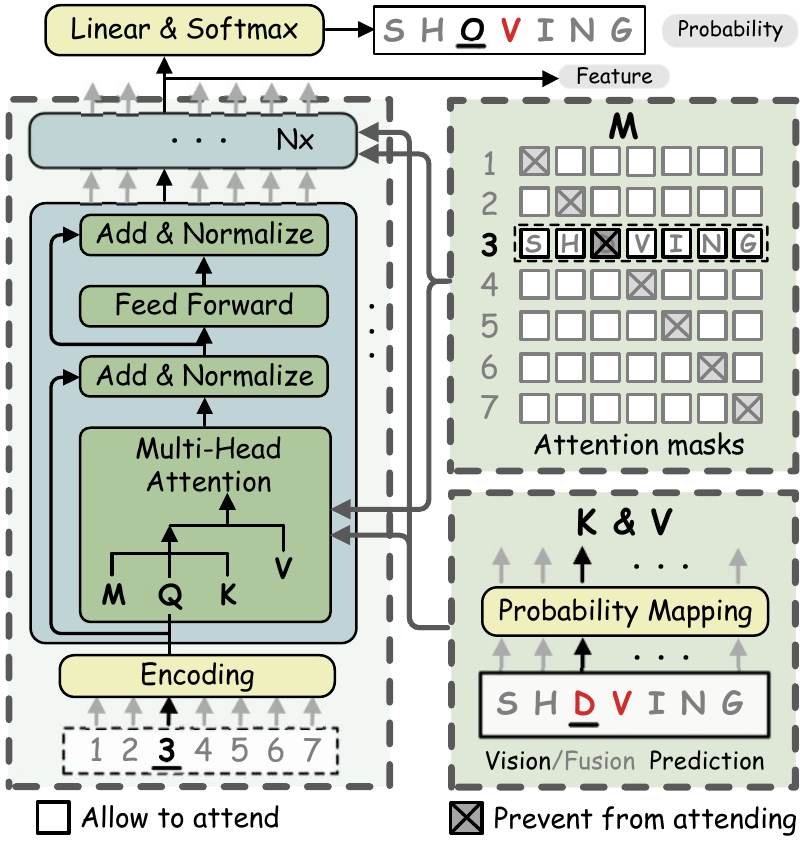}
      \caption{Architecture of language model (BCN).}
      \label{fig:label}
   \end{center}
   \vspace{-2em}
\end{figure}

Given a text string $\bm{y}=(y_1, \ldots, y_n)$ with text length $n$ and class number $c$, the conditional probability of $y_i$ for bidirectional and unidirectional models are $P(y_i|y_n,\dots,y_{i+1},y_{i-1},\dots,y_1)$ and $P(y_i|y_{i-1},\dots,y_1)$, respectively. From the perspective of information theory, available entropy of a bidirectional representation can be quantified as $H_{\bm{y}} = (n-1)\log{c}$. However, for a unidirectional representation the information is $\frac{1}{n}\sum^n_{i=1}{(i-1)\log{c}}=\frac{1}{2}H_{\bm{y}}$. Our insight is that previous methods typically use an ensemble model of two unidirectional models, which essentially are unidirectional representations. The unidirectional representation basically captures $\frac{1}{2}H_{\bm{y}}$ information, resulting in limited capability of feature abstraction compared with bidirectional counterpart.

Benefitting from the autonomous design in Section~\ref{sec:autonomous}, off-the-shelf NLP models with the ability of spelling correction can be transferred. A plausible way is utilizing the masked language model (MLM) in BERT~\cite{devlin2018bert} by replacing $y_i$ with token {\tt{[MASK]}}. However, we notice that this is unacceptable as MLM should be separately called $n$ times for each text instance, causing extreme low efficiency. Instead of masking the input characters, we propose BCN by specifying the attention masks.

Overall, the BCN is a variant of $L$-layers transformer decoder. Each layer of BCN is a series of multi-head attention and feed-forward network~\cite{vaswani2017attention} followed by residual connection~\cite{he2016deep} and layer normalization~\cite{ba2016layer}, as shown in Fig.~\ref{fig:label}. Different from vanilla Transformer, character vectors are fed into the multi-head attention blocks rather than the first layer of network. In addition, attention masks in multi-head attention are designed to prevent from ``seeing itself". Besides, no self-attention is applied in BCN to avoid leaking information across time steps. The attention operation inside multi-head blocks can be formalized as:
\begin{align}
\mathbf{M}_{ij} &= \begin{cases} 0, & i \neq j \\ -\infty, & i = j \end{cases}, \label{eq:att:mask} \\
\mathbf{K}_i &= \mathbf{V}_i = P(y_i) \mathbf{W}_l,  \\
\mathbf{F}_{mha} &= \text{softmax}(\frac{\mathbf{Q}\mathbf{K}^\mathsf{T}}{\sqrt{C}} + \mathbf{M})\mathbf{V},
\end{align}
where $\mathbf{Q} \in \mathbb{R}^{T \times C}$ is the positional encodings of character orders in the first layer and the outputs of the last layer otherwise. $\mathbf{K}, \mathbf{V} \in \mathbb{R}^{T \times C}$ are obtained from character probability $P(y_i) \in \mathbb{R}^{c}$, and $\mathbf{W}_l \in \mathbb{R}^{c \times C}$ is linear mapping matrix. $\mathbf{M} \in \mathbb{R}^{T \times T}$ is the matrix of attention masks which prevents from attending current character. After stacking BCN layers into deep architecture, the bidirectional representation $\mathbf{F}_{l}$ for text $\bm{y}$ is determined.

By specifying the attention masks in cloze fashion, BCN is able to learn more powerful bidirectional representation elegantly than the ensemble of unidirectional representation. Besides, benefitting from Transformer-like architecture, BCN can perform computation independently and parallelly. Also, it is more efficient than the ensemble models as only half of the computations and parameters are needed.

\subsubsection{Iterative Correction}
\label{sec:iterative}

The parallel-prediction of Transformer takes noise inputs which are typically approximations from visual prediction~\cite{yu2020towards} or visual feature~\cite{lyu20192d}. Concretely, as the example shown in Fig.~\ref{fig:framework} under bidirectional representation, the desired condition for $P(\text{``O"})$ is ``SH-WING". However, due to the blurred and occluded environments, the actual condition obtained from VM is ``SH-VING", in which ``V" becomes noise and harms the confidence of prediction. It tends to be more hostile for LM with increased error predictions in VM.

To cope with the problem of noise inputs, we propose iterative LM (illustrated in Fig.~\ref{fig:framework}). The LM is executed $M$ times repeatedly with different assignment for $\bm{y}$. For the first iteration, $\bm{y}_{i=1}$ is the probability prediction from VM. For the subsequent iterations, $\bm{y}_{i \ge 2}$ is the probability prediction from the fusion model (Section~\ref{sec:fusion}) in last iteration. By this way the LM is able to correct the vision prediction iteratively.

Another observation is that Transformer-based methods generally suffer from unaligned-length problem~\cite{yu2020towards}, which denotes that the Transformer is hard to correct the vision prediction if character number is unaligned with ground truth. The unaligned-length problem is caused by the inevitable implementation of padding mask which is fixed for filtering context outside text length. Our iterative LM can alleviate this problem as the visual feature and linguistic feature are fused several times, and thus the predicted text length is also refined gradually.

\subsection{Fusion}
\label{sec:fusion}

Conceptually, vision model trained on image and language model trained on text come from different modalities. To align visual feature and linguistic feature, we simply use the gated mechanism \cite{yu2020towards, yue2020robustscanner} for final decision:
\begin{align}
\mathbf{G} &= \sigma([\mathbf{F}_{v}, \mathbf{F}_{l}] \mathbf{W}_f), \\
\mathbf{F}_{f} &= \mathbf{G} \odot \mathbf{F}_{v} + (1 - \mathbf{G}) \odot \mathbf{F}_{l},
\end{align}
where $\mathbf{W}_f \in \mathbb{R}^{2C \times C}$ and $\mathbf{G} \in \mathbb{R}^{T \times C}$.

\subsection{Supervised Training}

ABINet is trained end-to-end using the following multi-task objectives:
\begin{align}
\mathcal{L} &= \lambda_v \mathcal{L}_v + \frac{\lambda_l}{M} \sum^M_{i=1}{\mathcal{L}^i_l} + \frac{1}{M} \sum^M_{i=1}{\mathcal{L}^i_f},
\label{eq:loss}
\end{align}
where $\mathcal{L}_v$, $\mathcal{L}_l$ and $\mathcal{L}_f$ are the cross entropy losses from $\mathbf{F}_{v}$, $\mathbf{F}_{l}$ and $\mathbf{F}_{f}$, respectively. Specifically, $\mathcal{L}^i_{l}$ and $\mathcal{L}^i_{f}$ are the losses at $i$-th iteration. $\lambda_v$ and $\lambda_l$ are balanced factors.



\subsection{Semi-supervised Ensemble Self-training}
\label{sec:semi-supervised}

To further explore the superiority of our iterative model, we propose a semi-supervised learning method based on self-training~\cite{xie2020self} with the ensemble of iterative predictions. The basic idea of self-training is first to generate pseudo labels by model itself, and then re-train the model using additional pseudo labels. Therefore, the key problem lies in constructing high-quality pseudo labels.

To filter the noise pseudo labels we propose the following methods: 1) minimum confidence of characters within a text instance is chosen as the text certainty. 2) Iterative predictions of each character are viewed as an ensemble to smooth the impact of noise labels. Therefore, we define the filtering function as follows:
\begin{align}
\begin{cases}
\mathcal{C} &= \min\limits_{1 \le t \le T} e^{\mathbb{E}[\log{P(y_t)}]} \\ 
P(y_t) &= \max\limits_{1 \le m \le M} P_m(y_t) 
\end{cases},
\label{eq:filter}
\end{align}
where $\mathcal{C}$ is the minimum \emph{certainty} of a text instance, $P_m(y_t)$ is probability distribution of $t$-th character at $m$-th iteration. The training procedure is depicted in Algorithm~\ref{alg:self-training}, where $Q$ is threshold. $B_l$, $B_u$ are training batches from labeled and unlabeled data. $N_{max}$ is the maximum number of training step and $N_{upl}$ is the step number for updating pseudo labels.

\begin{algorithm}[t]
   \scriptsize
   \caption{Ensemble Self-training}
   \begin{algorithmic}[1]
      \Require Labeled images $\mathcal{X}$ with labels $\mathcal{Y}$ and unlabeled images $\mathcal{U}$
      \State Train parameters $\theta_0$ of ABINet with $(\mathcal{X}$, $\mathcal{Y})$ using Equation~\ref{eq:loss}.
      \State Use $\theta_0$ to generate soft pseudo labels $\mathcal{V}$ for $\mathcal{U}$
      \State Get $(\mathcal{U}'$, $\mathcal{V}')$ by filtering $(\mathcal{U}$, $\mathcal{V})$ with $\mathcal{C}<Q$ (Equation~\ref{eq:filter})
      \For{$i = 1,\ldots, N_{max}$}
         \If{$i == N_{upl}$}
            \State Update $\mathcal{V}$ using $\theta_i$
            \State Get $(\mathcal{U}'$, $\mathcal{V}')$ by filtering $(\mathcal{U}$, $\mathcal{V})$ with $\mathcal{C}<Q$ (Equation~\ref{eq:filter})
         \EndIf
         \State Sample $B_l=(\mathcal{X}_{b}$, $\mathcal{Y}_{b}) \subsetneqq (\mathcal{X}$, $\mathcal{Y})$, $B_u=(\mathcal{U}'_{b}$, $\mathcal{V}'_{b}) \subsetneqq (\mathcal{U}'$, $\mathcal{V}')$
         \State Update $\theta_i$ with $B_l$, $B_u$ using Equation~\ref{eq:loss}.
      \EndFor
   \end{algorithmic}
   \label{alg:self-training}
\end{algorithm}

\section{Experiment}

\subsection{Datasets and Implementation Details}

Experiments are conducted following the setup of~\cite{yu2020towards} in the purpose of fair comparison. Concretely, the training datasets are two synthetic datasets MJSynth (MJ)~\cite{jaderberg2014synthetic,jaderberg2016reading} and SynthText (ST)~\cite{gupta2016synthetic}. Six standard benchmarks include ICDAR 2013 (IC13)~\cite{karatzas2013icdar}, ICDAR 2015 (IC15)~\cite{karatzas2015icdar}, IIIT 5K-Words (IIIT)~\cite{mishra2012scene}, Street View Text (SVT)~\cite{wang2011end}, Street View Text-Perspective (SVTP)~\cite{quy2013recognizing} and CUTE80 (CUTE)~\cite{risnumawan2014robust} are as the testing datasets. Details of these datasets can be found in the previous works~\cite{yu2020towards}. In addition, Uber-Text~\cite{Ying2017UberText} removing the labels is used as unlabeled dataset to evaluate the semi-supervised method.

The model dimension $C$ is set to 512 throughout. There are 4 layers in BCN with 8 attention heads each layer. Balanced factors $\lambda_v$, $\lambda_l$ are set to 1, 1 respectively. Images are directly resized to $32 \times 128$ with data augmentation such as geometry transformation (\ie, rotation, affine and perspective), image quality deterioration and color jitter, \etc. We use 4 NVIDIA 1080Ti GPUs to train our models with batch size 384. ADAM optimizer is adopted with the initial learning rate $1e^{-4}$, which is decayed to $1e^{-5}$ after 6 epochs.

\subsection{Ablation Study}
\subsubsection{Vision Model}

\begin{table}
   \begin{center}
   \caption{Ablation study of VM. Attn is the attention method and Trm Layer is the layer number of Transformer. SV, MV$_1$, MV$_2$ and LV are four VMs in different configurations.}
   \label{tab:vision}
	\resizebox{1.0\linewidth}{!}{
   \begin{tabular}{|c|c|c|c|c|c|c|c|c|}
      \hline
      Model & \multirow{2}{*}{Attn} & Trm & IC13 & SVT & IIIT & \multirow{2}{*}{Avg} & Params & Time\tablefootnote{Inference time is estimated using NVIDIA Tesla V100 by averaging 3 different trials.} \\
      Name &  & Layer & IC15 & SVTP & CUTE & & ($\times10^6$) & (ms) \\
      \hline
      SV & \multirow{2}{*}{parallel} & \multirow{2}{*}{2} &94.2& 89.6 & 93.7& \multirow{2}{*}{88.8} & \multirow{2}{*}{19.6} & \multirow{2}{*}{12.5} \\
      (small) & & & 80.6& 82.3& 85.1&  & & \\ 
      \hline
      MV$_1$ & \multirow{2}{*}{position} & \multirow{2}{*}{2} &93.6& 89.3 & 94.2& \multirow{2}{*}{89.0} & \multirow{2}{*}{20.4}  & \multirow{2}{*}{14.9} \\
      (middle) & & & 80.8& 83.1& 85.4&  &  & \\
      \hline
      MV$_2$ & \multirow{2}{*}{parallel} & \multirow{2}{*}{3} &94.5& 89.5 & 94.3& \multirow{2}{*}{89.4} & \multirow{2}{*}{22.8} & \multirow{2}{*}{14.8}  \\
      (middle) & & & 81.1& 83.7& \bf{86.8}&  & & \\  
      \hline
      LV & \multirow{2}{*}{position} & \multirow{2}{*}{3} &\bf{94.9}& \bf{90.4} & \bf{94.6}& \multirow{2}{*}{\bf{89.8}} & \multirow{2}{*}{23.5} & \multirow{2}{*}{16.7}  \\
      (large) & & & \bf{81.7} & \bf{84.2} & 86.5&  & & \\  
      \hline
   \end{tabular}}
   \end{center}
   \vspace{-1em}
\end{table}

Firstly, we discuss the performance of VM from two aspects: feature extraction and sequence modeling. Experiment results are recorded in Tab.~\ref{tab:vision}. The \emph{parallel} attention is a popular attention method~\cite{lyu20192d,yu2020towards}, and the proposed \emph{position} attention has a more powerful representation of key/value vectors. From the statistics we can conclude: 1) simply upgrading the VM will result in great gains in accuracy but at the cost of parameter and speed. 2) To upgrade the VM, we can use the position attention in feature extraction and a deeper transformer in sequence modeling.

\subsubsection{Language Model}

\begin{table}
   \begin{center}
   \caption{Ablation study of autonomous strategy. PVM is pre-training VM on MJ and ST in supervised way. PLM$_{in}$ is pre-training LM using text on MJ and ST in self-supervised way. PLM$_{out}$ is pre-training LM on WikiText-103~\cite{merity2016pointer} in self-supervised way. AGF means allowing gradient flow between VM and LM.}
   \label{tab:autonomous}
	\resizebox{0.9\linewidth}{!}{
   \begin{tabular}{|c|c|c|c|c|c|c|c|}
      \hline
      \multirow{2}{*}{PVM} & \multirow{2}{*}{PLM$_{in}$} & \multirow{2}{*}{PLM$_{out}$} & \multirow{2}{*}{AGF} & IC13 & SVT & IIIT & \multirow{2}{*}{Avg}  \\
       & & & & IC15 & SVTP & CUTE &  \\
      \hline
      \multirow{2}{*}{-} & \multirow{2}{*}{-} & \multirow{2}{*}{-} & \multirow{2}{*}{-} & 96.7 & 93.4 & 95.7 & \multirow{2}{*}{91.7}  \\
      & & & & 84.5 & 86.8 & 86.8&   \\ 
      \hline
      \multirow{2}{*}{\ding{51}} & \multirow{2}{*}{-} & \multirow{2}{*}{-} & \multirow{2}{*}{-} & 97.0 & 93.0 & 96.3 & \multirow{2}{*}{92.3}  \\
       & & & & 85.0 & 88.5 & 89.2&  \\ 
       \hline
      \multirow{2}{*}{-} & \multirow{2}{*}{\ding{51}} & \multirow{2}{*}{-} & \multirow{2}{*}{-} & 97.1 & \bf{93.8} & 95.5 & \multirow{2}{*}{91.6}  \\
       & & & & 83.6 & 88.1 & 86.8&  \\ 
      \hline
      \multirow{2}{*}{\ding{51}} & \multirow{2}{*}{\ding{51}} & \multirow{2}{*}{-} & \multirow{2}{*}{-} & \bf{97.2} & 93.5 & 96.3 & \multirow{2}{*}{92.3}  \\
      & & & & 84.9 & \bf{89.0} & 88.5&  \\ 
      \hline
      \multirow{2}{*}{\ding{51}} & \multirow{2}{*}{-} & \multirow{2}{*}{\ding{51}} & \multirow{2}{*}{-} & 97.0 & 93.7 & \bf{96.5} & \multirow{2}{*}{\bf{92.5}}  \\
      & & & & \bf{85.3} & 88.5 & \bf{89.6}&  \\ 
      \hline
      \multirow{2}{*}{\ding{51}} & \multirow{2}{*}{-} & \multirow{2}{*}{-} & \multirow{2}{*}{\ding{51}} & 96.7 & 92.6 & 95.7& \multirow{2}{*}{91.4}  \\
      & & & & 83.3 & 86.5 & 88.5&  \\ 
      \hline
   \end{tabular}}
   \end{center}
   \vspace{-2em}
\end{table}

\paragraph{Autonomous Strategy.} To analyze the autonomous models, we adopt the LV and BCN as VM and LM respectively. From the results in Tab.~\ref{tab:autonomous} we can observe: 1) pre-training VM is useful which boosts the accuracy about $0.6\%$-$0.7\%$ on average; 2) the benefit of pre-training LM on the training datasets (\ie, MJ and ST) is negligible; 3) while pre-training LM from an additional unlabeled dataset (\eg, WikiText-103) is helpful even when the base model is in high accuracy. The above observations suggest that it is useful for STR to pre-train both VM and LM. Pre-training LM on additional unlabeled datasets is more effective than on training datasets since the limited text diversity and biased data distribution are unable to facilitate the learning of a well-performed LM. Also, pre-training LM on unlabeled datasets is cheap since additional data is available easily.

Besides, by allowing gradient flow (AGF) between VM and LM, the performance decreases $0.9\%$ on average (Tab.~\ref{tab:autonomous}. We also notice that the training loss of AGF reduces sharply to a lower value. This indicates that overfitting occurs in LM as the VM helps to cheat in training, which might also happen in implicitly language modeling. Therefore it is crucial to enforce LM to learn independently by BGF. We note that SRN~\cite{yu2020towards} uses \emph{argmax} operation after VM, which is intrinsically a special case of BGF since \emph{argmax} is non-differentiable. Another advantage is that the autonomous strategy makes the models a better interpretability, since we can have a deep insight into the performance of LM (\eg, Tab.~\ref{tab:spelling_correction}), which is infeasible in implicitly language modeling.

\begin{table}
   \begin{center}
   \caption{Ablation study of bidirectional representation.}
   \label{tab:bidirectional}
	\resizebox{0.9\linewidth}{!}{
   \begin{tabular}{|c|c|c|c|c|c|c|c|}
      \hline
      \multirow{2}{*}{Vision} & \multirow{2}{*}{Language} & IC13 & SVT & IIIT & \multirow{2}{*}{Avg} & Params & Time \\
       &  & IC15 & SVTP & CUTE & & ($\times10^6$) & (ms) \\
      \hline
       & \multirow{2}{*}{SRN-U} & 96.0 & 90.3 & 94.9& \multirow{2}{*}{90.2} & \multirow{2}{*}{32.8} & \multirow{2}{*}{19.1} \\
       & & 81.9 & 86.0 & 85.4&  & & \\ 
      \cline{2-8}
      \multirow{2}{*}{SV} & \multirow{2}{*}{SRN} & 96.3 & 90.9 & 95.0 & \multirow{2}{*}{90.6} & \multirow{2}{*}{45.4} & \multirow{2}{*}{24.2} \\
      & & 82.6 & \bf{86.4} & 87.5&  & & \\ 
      \cline{2-8}
      & \multirow{2}{*}{BCN} & \bf{96.7} & \bf{91.7} & \bf{95.3} & \multirow{2}{*}{\bf{91.0}} & \multirow{2}{*}{32.8} & \multirow{2}{*}{19.5} \\
      & & \bf{83.1} & 86.2 & \bf{88.9}&  & & \\ 
      \hline
      \hline
      & \multirow{2}{*}{SRN-U} & 96.0 & 91.2 & 96.2 & \multirow{2}{*}{91.5} & \multirow{2}{*}{36.7} & \multirow{2}{*}{22.1} \\
      & & 84.0 & 86.8 & 87.8&  & & \\ 
      \cline{2-8}
      \multirow{2}{*}{LV} & \multirow{2}{*}{SRN} & 96.8 & 92.3 & \bf{96.3} & \multirow{2}{*}{91.9} & \multirow{2}{*}{49.3} & \multirow{2}{*}{26.9} \\
      & & 84.2 & 87.9 & 88.2&  & & \\ 
      \cline{2-8}
      & \multirow{2}{*}{BCN} & \bf{97.0} & \bf{93.0} & \bf{96.3} & \multirow{2}{*}{\textbf{92.3}} & \multirow{2}{*}{36.7} & \multirow{2}{*}{22} \\
      & & \bf{85.0} & \bf{88.5} & \bf{89.2}&  & & \\ 
      \hline
   \end{tabular}}
   \end{center}
   \vspace{-1em}
\end{table}

\begin{table}
   \begin{center}
   \caption{Top-5 accuracy of LMs in text spelling correction.}
   \resizebox{0.85\linewidth}{!}{
   \begin{tabular}{|c|c|c|}
      \hline
      Language Model & Character Accuracy & Word Accuracy \\
      \hline
      SRN & 78.3 & 27.6 \\
      \hline
      BCN & \bf{82.8} & \bf{41.9} \\
      \hline
   \end{tabular}}
   \label{tab:spelling_correction}
   \end{center}
   \vspace{-2em}
\end{table}

\vspace{-1em}

\paragraph{Bidirectional Representation.} As the BCN is a variant of Transformer, we compare BCN with its counterpart SRN. The Transformer-based SRN~\cite{yu2020towards} shows superior performance which is an ensemble of unidirectional representation. For fair comparison, experiments are conducted with the same conditions except the networks. We use SV and LV as the VMs to validate the effectiveness at different accuracy levels. As depicted in Tab.~\ref{tab:bidirectional}, though BCN has similar parameters and inference speed as the unidirectional version of SRN (SRN-U), it achieves competitive advantage in accuracy under different VMs. Besides, compared with the bidirectional SRN in ensemble, BCN shows better performance especially on challenging datasets such as IC15 and CUTE. Also, ABINet equipped with BCN is about $20\%$-$25\%$ faster than SRN, which is practical for large-scale tasks.

Section~\ref{sec:autonomous} has argued that the LMs can be viewed as independent units to estimate the probability distribution of spelling correction, and thus we conduct experiments from this view. The training set is the text from MJ and ST. To simulate spelling errors, the testing set is 20000 items which are chosen randomly, where we add or remove a character for $20\%$ text, replace a character for $60\%$ text and keep the rest of the text unchangeable. From the results in Tab.~\ref{tab:spelling_correction}, we can see BCN outperforms SRN by $4.5\%$ character accuracy and $14.3\%$ word accuracy, which indicates that BCN has a more powerful ability in character-level language modeling.

\begin{figure}
   \begin{center}
      \includegraphics[width=0.48\textwidth]{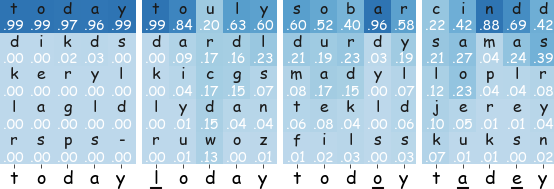}
      \caption{Visualization of top-5 probability in BCN.}
      \label{fig:visual_topk}
   \end{center}
   \vspace{-1.0em}
\end{figure}

To better understand how BCN works inside ABINet, we visualize the top-5 probability in Fig.~\ref{fig:visual_topk}, which takes ``today" as an example. On the one hand, as ``today" is a string with semantic information, taking ``-oday" and ``tod-y" as inputs, BCN can predict ``t" and ``a" with high confidence and contribute to final fusion predictions. On the other hand, as error characters ``l" and ``o" are noise for the rest predictions, BCN becomes less confident and has little impact to final predictions. Besides, if there are multiple error characters, it is hard for BCN to restore correct text due to lacking of enough context.

\begin{table}
   \begin{center}
   \caption{Ablation study of iterative correction.}
   \label{tab:iterative}
	\resizebox{0.9\linewidth}{!}{
   \begin{tabular}{|c|c|c|c|c|c|c|c|}
      \hline
      \multirow{2}{*}{Model} & Iteration & IC13 & SVT & IIIT & \multirow{2}{*}{Avg} & Params & Time \\
       & Number & IC15 & SVTP & CUTE & & ($\times10^6$) & (ms) \\
      \hline
      \multirow{2}{*}{SV} & \multirow{2}{*}{1} & 96.7 & 91.7 & 95.3 & \multirow{2}{*}{91.0} & \multirow{2}{*}{32.8} & \multirow{2}{*}{19.5} \\
      &  & 83.1 & 86.2  & 88.9 &  & & \\ 
      \cline{2-8}
      \multirow{2}{*}{+} & \multirow{2}{*}{2} & \bf{97.2} & 91.8 & \bf{95.4} & \multirow{2}{*}{91.2} & \multirow{2}{*}{32.8} & \multirow{2}{*}{24.5} \\
      &  & 83.3 & 86.4 & 89.2 &  & & \\ 
      \cline{2-8}
      \multirow{2}{*}{BCN} & \multirow{2}{*}{3}  & 97.1 & \bf{93.0} & \bf{95.4} & \multirow{2}{*}{\bf{91.4}} & \multirow{2}{*}{32.8} & \multirow{2}{*}{31.6} \\
      & & \bf{83.4} & \bf{86.7}  & \bf{89.6}&  & & \\ 
      \hline
      \hline
      \multirow{2}{*}{LV} & \multirow{2}{*}{1} & 97.0 & 93.0 & 96.3 & \multirow{2}{*}{92.3} & \multirow{2}{*}{36.7} & \multirow{2}{*}{22} \\
      &  & 85.0 & 88.5 & 89.2&  & & \\ 
      \cline{2-8}
      \multirow{2}{*}{+} & \multirow{2}{*}{2} & 97.1 & 93.4 & 96.3 & \multirow{2}{*}{92.4} & \multirow{2}{*}{36.7} & \multirow{2}{*}{27.3} \\
      &  & 85.2 & 88.7 & \bf{89.6}&  & & \\ 
      \cline{2-8}
      \multirow{2}{*}{BCN} & \multirow{2}{*}{3} & \bf{97.3} & \bf{94.0} & \bf{96.4} & \multirow{2}{*}{\bf{92.6}} & \multirow{2}{*}{36.7} & \multirow{2}{*}{33.9} \\
      &  & \bf{85.5} & \bf{89.1}  & 89.2 &  & & \\ 
      \hline
   \end{tabular}}
   \end{center}
   \vspace{-2.0em}
\end{table}

\begin{table*}[htp]
   \vspace{0em}
   \begin{center}
   \caption{Accuracy comparison with other methods.}
   \label{tab:benchmark}
   \resizebox{0.85\linewidth}{!}{
   \begin{tabular}{|c|l|c|c|c|c|c|c|c|c|c|}
   \hline
   \multirow{2}{*}{} & \multirow{2}{*}{Methods} & Labeled & Unlabeled & \multicolumn{3}{|c|}{Regular Text}  & \multicolumn{3}{|c|}{Irregular Text} \\
   \cline{4-10}
   & & Datasets & Datasets & IC13 & SVT & IIIT & IC15 & SVTP & CUTE \\
   \hline
   \multirow{7}{*}{\rotatebox{90}{SOTA methods}} & 2019~Lyu~\etal~\cite{lyu20192d}~(Parallel) & MJ+ST & - & 92.7 & 90.1 & 94.0 & 76.3 & 82.3& 86.8 \\
   & 2019~Liao~\etal~\cite{liao2019mask}~(SAM) & MJ+ST & - & 95.3 & 90.6 & 93.9 & 77.3 & 82.2 & 87.8 \\
   & 2020~Qiao~\etal~\cite{qiao2020seed}~(SE-ASTER) & MJ+ST & - & 92.8 & 89.6 & 93.8 & 80.0 & 81.4 & 83.6 \\
   & 2020~Wan~\etal~\cite{wan2019textscanner}~(Textscanner) & MJ+ST & - & 92.9 & 90.1 & 93.9 & 79.4 & 84.3 & 83.3 \\
   & 2020~Wang~\etal~\cite{wang2020decoupled}~(DAN) & MJ+ST & - & 93.9 & 89.2 & 94.3 & 74.5 & 80.0 & 84.4 \\
   & 2020~Yue~\etal~\cite{yue2020robustscanner}~(RobustScanner) & MJ+ST & - & 94.8 & 88.1 & 95.3 & 77.1 & 79.5 & \bf{90.3}\\
   & 2020~Yu~\etal~\cite{yu2020towards} (SRN) & MJ+ST & - & 95.5 & 91.5 & 94.8 & 82.7 &85.1 &87.8\\
   \hline
   \multirow{6}{*}{\rotatebox{90}{Ours}} & SRN-SV (Reproduced) &  MJ+ST & - & 96.3 & 90.9 & 95.0 & 82.6 & 86.4 & 87.5 \\
   & ABINet-SV & MJ+ST & - & \bf{96.8} & \bf{93.2} & \bf{95.4} & \bf{84.0} &  \bf{87.0} & 88.9 \\  
   \cline{2-10}
   & SRN-LV (Reproduced)  &  MJ+ST & - & 96.8 & 92.3 & \bf{96.3} & 84.2 & 87.9 & 88.2 \\
   & ABINet-LV & MJ+ST & - & \bf{97.4} & \bf{93.5} & 96.2 & \bf{86.0} & \bf{89.3}  & \bf{89.2} \\
   \cline{2-10}
   & ABINet-LV${_{st}}$ & MJ+ST & Uber-Text & 97.3 & 94.9 & 96.8 & \bf{87.4} & \bf{90.1}  & 93.4 \\  
   & ABINet-LV${_{est}}$ & MJ+ST & Uber-Text & \bf{97.7} & \bf{95.5} & \bf{97.2} & 86.9 & 89.9  & \bf{94.1} \\  
   \hline
   \end{tabular}}
   \end{center}
   \vspace{-2em}
 \end{table*}


\vspace{-1em}

\paragraph{Iterative Correction.} We apply SV and LV again with BCN to demonstrate the performance of iterative correction from different levels. Experiment results are given in Tab.~\ref{tab:iterative}, where the iteration numbers are set to 1, 2 and 3 both in training and testing. As can be seen from the results, iterating the BCN 3 times can respectively boost the accuracy by $0.4\%$, $0.3\%$ on average. Specifically, there are little gains on IIIT which is a relatively easy dataset with clear character appearance. However, when it comes to other harder datasets such as IC15, SVT and SVTP, the iterative correction steadily increases the accuracy and achieves up to $1.3\%$ and $1.0\%$ improvement on SVT for SV and LV respectively. It is also noted that the inference time increases linearly with the iteration number.

\begin{figure}
   \begin{center}
      \includegraphics[width=0.48\textwidth]{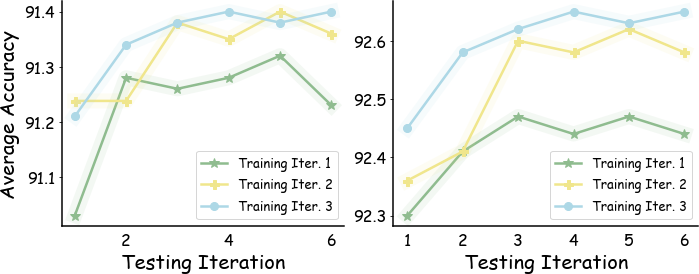}
      \caption{Accuracy of iterating BCN in training and testing.}
      \label{fig:iteration}
   \end{center}
   \vspace{-0em}
\end{figure}

We further explore the difference of iteration between training and testing. The fluctuation of average accuracy in Fig.~\ref{fig:iteration} suggests that: 1) directly applying iterative correction in testing also works well; 2) while iterating in training is beneficial since it provides additional training samples for LM; 3) the accuracy reaches a saturated state when iterating the model more than 3 times, and therefore a big iteration number is unnecessary.

\begin{figure}
   \vspace{-1.5em}
   \begin{center}
      \includegraphics[width=0.5\textwidth]{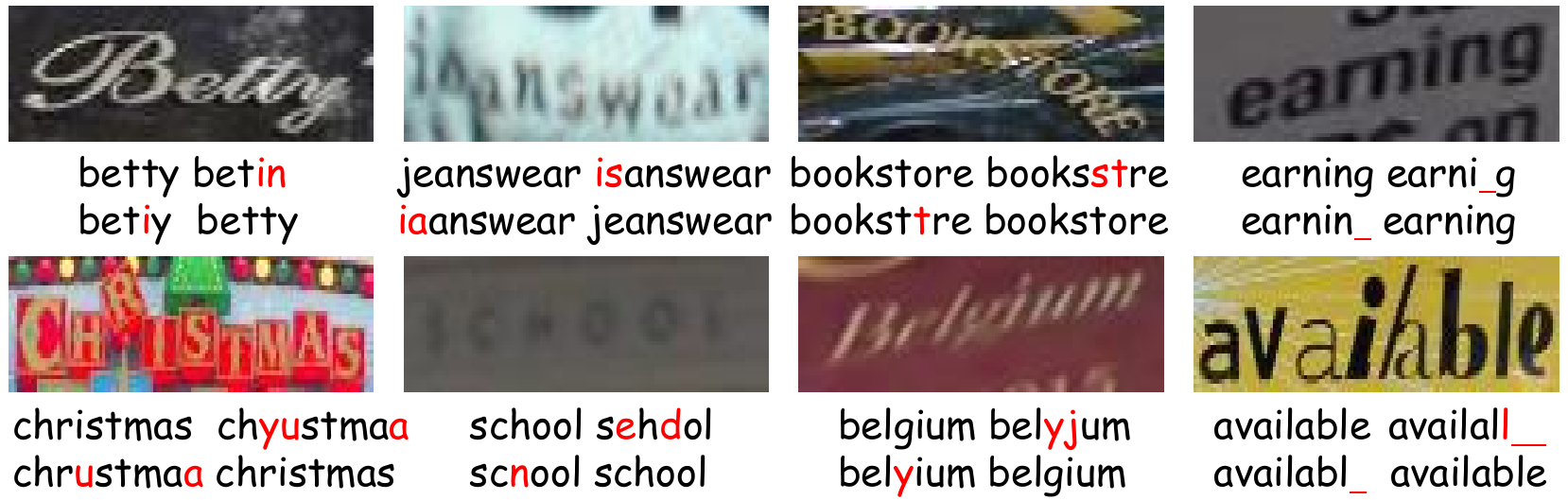}
      \caption{Successful examples using iterative correction. Text strings are ground truth, vision prediction, fusion prediction without iterative correction and with iterative correction respectively from left to right and top to bottom.}
      \label{fig:iteration_img}
   \end{center}
   \vspace{-2.5em}
\end{figure}

To have a comprehensive cognition about iterative correction, we visualize the intermediate predictions in Fig.~\ref{fig:iteration_img}. Typically, the vision predictions can be revised approaching to ground truth while remain errors in some cases. After multiple iterations, the predictions can be corrected finally. Besides, we also observe that iterative correction is able to alleviate the unaligned-length problem, as shown in the last column in Fig.~\ref{fig:iteration_img}. 

From the ablation study we can conclude: 1) the bidirectional BCN is a powerful LM which can effectively improve the performance both in accuracy and speed. 2) By further equipping BCN with iterative correction, the noise input problem can be alleviated, which is recommended to deal with challenging examples such as low-quality images at the expense of incremental computations.

\subsection{Comparisons with State-of-the-Arts}

 Generally, it is not an easy job to fairly compare with other methods directly using the reported statistics~\cite{baek2019wrong}, as differences might exist in backbone (\ie, CNN structure and parameters), data processing (\ie, images rectification and data augmentation) and training tricks, \etc. To strictly perform fair comparison, we reproduce the SOTA algorithm SRN which shares the same experiment configuration with ABINet, as presented in Tab.~\ref{tab:benchmark}. The two reimplemented SRN-SV and SRN-LV are slightly different from the reported model by replacing VMs, removing the side-effect of multi-scales training, applying decayed learning rate, etc. Note that SRN-SV performs somewhat better than SRN due to the above tricks. As can be seen from the comparison, our ABINet-SV outperforms SRN-SV with $0.5\%$, $2.3\%$, $0.4\%$, $1.4\%$, $0.6\%$, $1.4\%$ on IC13, SVT, IIIT, IC15, SVTP and CUTE datasets respectively. Also, the ABINet-LV with a more strong VM achieve an improvement of $0.6\%$, $1.2\%$, $1.8\%$, $1.4\%$, $1.0\%$ on IC13, SVT, IC15, SVTP and CUTE benchmarks over its counterpart.
 
 Compared with recent SOTA works that are trained on MJ and ST, ABINet also shows impressive performance~(Tab.~\ref{tab:benchmark}). Especially, ABINet has prominent superiority on SVT, SVTP and IC15 as these datasets contain a large amount of low-quality images such as noise and blurred images, which the VM is not able to confidently recognize. Besides, we also find that images with unusual-font and irregular text can be successfully recognized as the linguistic information acts as an important complement to visual feature. Therefore ABINet can obtain second best result on CUTE even without image rectification.

\subsection{Semi-Supervised Training}

\begin{figure}
   \begin{center}
      \includegraphics[width=0.5\textwidth]{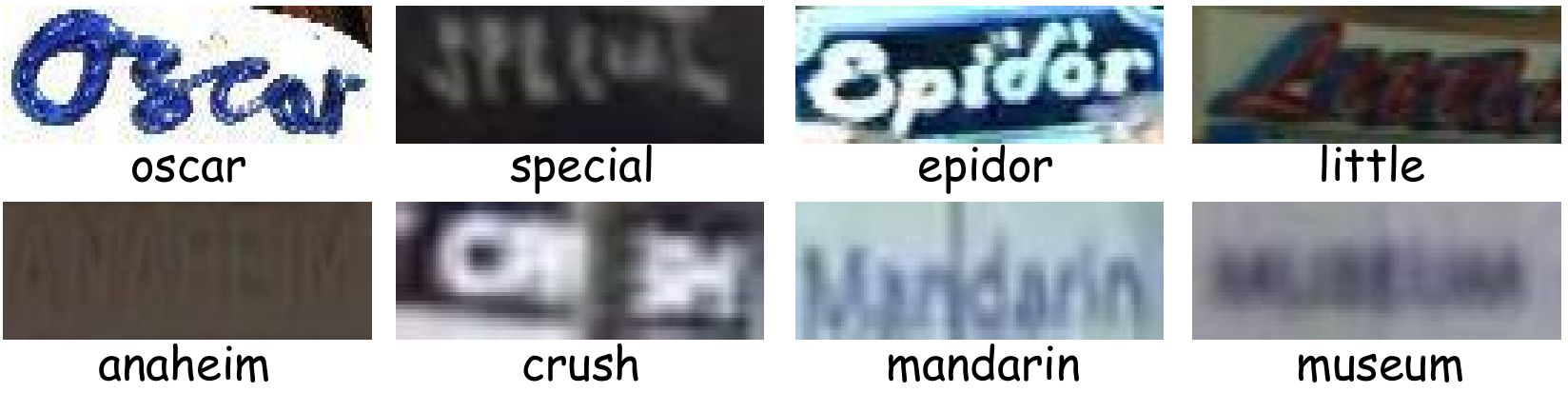}
      \caption{\textls[-1]{Hard examples successfully recognized by ABINet-LV${_{est}}$.}}
      \label{fig:semi_img}
   \end{center}
   \vspace{-2.5em}
\end{figure}

To further push the boundary of accurate reading, we explore a semi-supervised method which utilizes MJ and ST as the labeled datasets and Uber-Text as the unlabeled dataset. The threshold $Q$ in Section~\ref{sec:semi-supervised} is set to 0.9, and the batch size of $B_l$ and $B_u$ are 256 and 128 respectively. Experiment results in Tab.~\ref{tab:benchmark} show that the proposed self-training method ABINet-LV${_{st}}$ can easily outperform ABINet-LV on all benchmark datasets. Besides, the ensemble self-training ABINet-LV${_{est}}$ shows a more stable performance by improving the efficiency of data utilization. Observing the boosted results we find that hard examples with scarce fonts and blurred appearance can also be recognized frequently~(Fig.~\ref{fig:semi_img}), which suggests that exploring the semi-/unsupervised learning methods is a promising direction for scene text recognition.

\vspace{-1em}
\section{Conclusion}

In this paper, we propose ABINet which explores effective approaches for utilizing linguistic knowledge in scene text recognition. The ABINet is 1) autonomous that improves the ability of language model by enforcing learning explicitly; 2) bidirectional that learns text representation by jointly conditioning on character context at both sides; and 3) iterative that corrects the prediction progressively to alleviate the impact of noise input. Based on the ABINet we further propose an ensemble self-training method for semi-supervised learning. Experiment results on standard benchmarks demonstrate the superiority of ABINet especially on low-quality images. In addition, we also claim that exploiting unlabeled data is possible and promising for achieving human-level recognition.



{\small
\bibliographystyle{ieee_fullname}
\bibliography{egbib}
}

\end{document}